# Shallow Unorganized Neural Networks Using Smart Neuron Model for Visual Perception

**Richard Jiang[1] and Danny Crookes[2]**
[1]Computing and Communication, Lancaster University, Lancaster, UK
[2]Computer Science, Queen's University Belfast, Belfast, UK

Corresponding author: Richard Jiang (e-mail: r.jiang2@lancaster.ac.uk)

This work was supported in part by the EPSRC grant (EP/P009727/1).

**ABSTRACT** The recent success of Deep Neural Networks (DNNs) has revealed the significant capability of neural computing in many challenging applications. Although DNNs are derived from emulating biological neurons, there still exist doubts over whether or not DNNs are the final and best model to emulate the mechanism of human intelligence. In particular, there are two discrepancies between computational DNN models and the observed facts of biological neurons. First, human neurons are interconnected randomly, while DNNs need carefully-designed architectures to work properly. Second, human neurons usually have a long spiking latency (~100*ms*) which implies that not many layers can be involved in making a decision, while DNNs could have hundreds of layers to guarantee high accuracy. In this paper, we propose a new computational model, namely *shallow unorganized neural networks* (SUNNs), in contrast to ANNs/DNNs. The proposed SUNNs differ from standard ANNs or DNNs in three fundamental aspects: 1) SUNNs are based on an adaptive neuron cell model, *Smart Neurons*, that allows each artificial neuron cell to adaptively respond to its inputs rather than carrying out a fixed weighted-sum operation like the classic neuron model in ANNs/DNNs; 2) SUNNs can cope with computational tasks with very shallow architectures; 3) SUNNs have a natural topology with random interconnections, as the human brain does, and as proposed by Turing's B-type unorganized machines. We implemented the proposed SUNN architecture and tested it on a number of unsupervised early stage visual perception tasks. Surprisingly, such simple shallow architectures achieved very good results in our experiments. The success of our new computational model makes it the first workable example of Turing's B-Type unorganized machine that can achieve comparable or better performance against the state-of-the-art algorithms.

**INDEX TERMS**: Unorganized neural networks, Turing's type-B unorganized machine, smart neuron model, early vision, unsupervised visual processing

## I. INTRODUCTION

THE emulation of human intelligence has been a long-term driving force for the birth and growth of Artificial Intelligence (AI) since Alan Turing initially proposed its concept in the 1930s [1, 2]. The simple weighted-sum neuron model [2, 3] as shown in Fig.1-a) and b) is the abstraction of biological neurons with dendrites which has served as the fundamental building block for artificial neural networks (ANNs) since the 1960s. ANNs are well designed architectures of multiple layers of such artificial neurons, which are also called Multi-Layer Perceptrons (MLPs) [3].

Most MLPs have a structure of three layers, which is very popular among AI community in 1980s and 1990s [2, 3]. Though MLPs have been seen useful in many applications, their performance was overtaken by other machine learning techniques, such as Support Vector Machine [3], and the use of ANNs fell off in the 1990s and early 2000s. The resurrection of ANNs came with the birth of Deep Learning, where a sophisticated backpropagation algorithm proposed by Hinton *et al* [4~6] enabled various deep layered architectures.

It has been reported some deep neural networks (DNNs) could have more than 1000 layers of neurons [7], making it a tremendous task to train these DNNs. DNNs benefit from fine-trained parameters at double precision, nicely designed deep architectures, and subtle partitions of learning subspaces over many layers. As a consequence DNNs have outperformed human brains in many challenging tasks, such as DeepMind's AlphaGo [8] and video game play [9].

While deep learning has acieved great success in many applications [4~17], there are critics [18~22] of the bio-plausibility of DNNs. First, though back propagation is very effective, there is no such biological process to match such differentiation-based calculation and tuning at double-precision. How human intelligence can cope with complicated tasks is not yet known, and neuron scientists are still looking for final answers.

Secondly, DNNs such as convolutional neural networks (CNNs) are based on carefully-defined architectures over neighboring pixels or signals. However, biological neurons are more likely randomly interconnected. This raises the curious question [18, 19] as to why such randomly interconnected networks can work as well as, or even outperform, carefully-designed architectures.

Thirdly, as pointed out by Thorpe *et al* [23~25], the response time in the human brain is roughly 100~300ms, equal to about 1-3 spiking intervals. This implies that some human decisions are made by a very shallow layered structure within 1~3 spiking layers. Hence, biological neural computation should not rely on propagation of spiking signals over too many layers, certainly not hundreds of layers, though DNNs/ANNs can do so with no biophysical restriction.

Fourthly, in biological neurons, the input synaptic weights [19, 25, 41, 52] cannot be negative, and it is unlikely they operate at double precision. The ideal conditions assumed in the ANN neuron model do not match with the facts of biological neurons. Instead, biological neural networks (BNNs) [25] are more like a fault-tolerant system with fuzzy values that may not be very accurate numerically.

Michael I. Jordan [18] has summarized the challenges facing DNNs:

*"we still don't know how neurons learn. Is it actually just a small change in the synaptic weight that's responsible for learning? …we have precious little idea how learning (in brain) is actually taking place."*

This comment was echoed by other top AI researchers recently as well [19~22]. Uncovering the secrets of human intelligence has been a major motivation in AI research; however, there exist fundamental questions which are unanswered by current state-of-the-art ANNs and DNNs.

In this paper, we target the above critical issues. We reexamine the biological plausibility of neural networks, and propose a new artificial neural network model, *Shallow Unorganized Neural Networks* (SUNNs), which can better fit with the observed biological facts of human brains. Our proposed new neural computing model differs from standard ANNs or DNNs in three fundamental aspects,

1) Our computing model is based on a new neuron model, namely *Smart Neurons*. Unlike ANNs using a simple weighted sum model, our smart neuron model (SNM) can discriminatively select signals, and the overall intelligence of SUNNs comes from the collective intelligence of single smart neurons.
2) Unlike most ANNs and DNNs which have a well-defined interconnected architecture, our SUNNs are unorganized and all interconnections among neurons are generated randomly, which is a practical case of Turing's type-B unorganized machine [26, 27].
3) Our SUNNs have a very shallow architecture. In this work, we propose a dual-layer architecture for some unsupervised visual tasks (early vision tasks), where outputs can be generated simply by only one spiking latency.

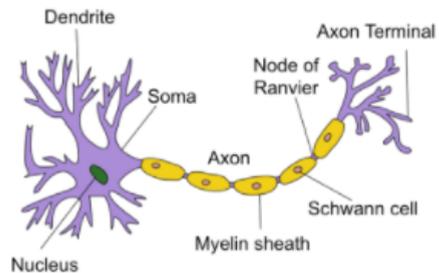

a) The typical structure of a biological neuron

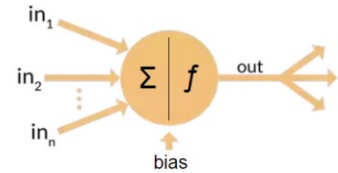

b) Simplified neural computing model in ANN

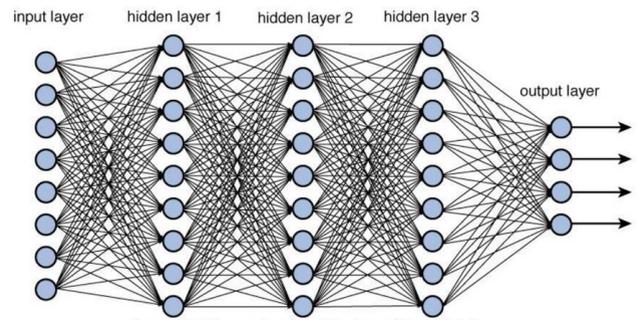

c) Carefully-designed multi-layered neural network

Figure 1. Neurons and standard ANNs and DNNs

We implemented our proposed SUNNs and tested them in unsupervised early vision tasks. Surprisingly, without carefully-designed architectures, fine-tuned parameters and deeper layers, our SUNNs demonstrated very good performance in these challenging tasks.

To make it clear, our SUNN model is not a biophysical model like Spiking Neural Networks. Instead, it is simply another type of ANNs for pure computational purpose, though its basic mechanism is different from the standard ANNs and includes more bio-plausible aspects in its smart neuron model as well as its randomly-interconnected architecture.

John McCarthy, who coined the term Artificial Intelligence in 1955, commented that [28],

*"As soon as it works, no one calls it AI anymore."*

Turing's type-B machine [26] has been mostly considered a theoretical concept in AI. To the best of our knowledge, our work gives the first example of such Type-B machines that can really work on complicated AI tasks such as visual perception, and which can achieve similar or better performance as state-of-the-art algorithms.

In the following sections, section II presents the motivation of this work based on analysis of biological facts; section III attempts to establish a theoretical foundation on how to make a randomly interconnected network work for machine learning tasks; and section IV presents our proposed Smart Neuron Model as the building blocks for SUNNs. Section V gives a specific case of SUNNs with a dual-layer structure, and section VI validates these SUNNs on several interesting unsupervised visual tests, including edge extraction, object pop-out, and image layer separation. Section VII carries out a deeper discussion of wider issues. Section VIII concludes the paper.

## II. MOTIVATION OF THIS WORK

### A. OVERVIEW OF ARTIFICIAL NEURAL NETWORKS

The past decade has witnessed remarkable progress in DNNs. DNN-based systems now outperform expert humans at Atari video games[9], the ancient board game Go[8], and high-stakes matches of heads-up poker [10, 11]. DNNs can also translate between multiple languages, produce handwriting [12] and speech [13] indistinguishable from those of humans, and even reformat your holiday snaps in the style of Van Gogh [14] masterpieces.

These advances are attributed to the success of DNNs, which is often considered a derivation from neuroscience. Simulating human intelligence played a key role in driving the research in this direction, originally motivated by a desire to understand how the human brain works. In fact, throughout the past half century, much of the key work developing neural networks took place not in mathematics or physics labs, but in psychology and neurophysiology departments [2, 19].

Active research is ongoing by those who are aware how different these deep networks are from what neural scientists have discovered decades ago in mammalian brain tissue. And there are more differentiators being discovered today as learning circuitry and neuro-chemistry in the brain is investigated from the genomic perspective.

There are apparent differences between DNNs and BNNs. For example, as shown in Fig.2, BNNs have much more complex topological structure featured by random interconnections. In contrast, ANNs and DNNs rely heavily on well-designed architectures as shown in Fig.1-c). Such random interconnections were a feature of Turing's B-type unorganized machine [26], which is characterised by its unorganized architecture. It is the most apparent difference between BNNs and ANNs/DNNs, while random interconnection is a biological nature of human brain.

In ANNs, as shown in Fig.1, a neuron cell is simply modelled as a weighted sum of inputs, in which weights are fixed once trained. In real biological circuitry, due to the topological complexity, there could be cross-talk effects between interconnections and leaky currents may play important role in the origin of intelligence. As we know, mammalian brains have dozens of neuro-transmitter and neuro-regulation compounds that have regional effects on circuitry [25]. The role of chemistry may be essential to learning social and reproductive behaviour that interrelates DNA information with propagation, linking in complex ways learning at the level of an ecosystem and the brain. Furthermore, long term and short term learning divides the brain's learning into two distinct capabilities. Indeed, living cells have many substructures and it is known that several types have complex relationships with signal transmission in neurons.

As a consequence of such differences between BNNs and ANNs, human eyes can easily identify patterns in the presence of slight noise, as shown in Fig.3, while a well-trained DNN-based classifier can be fooled by added noise into mistaking a panda for a gibbon [20]. A well-defined DNN architecture is sensitive to values and needs to be tuned at double precision, while fault-tolerant BNN systems can handle minor changes and still identify the main content.

### B. CHALLENGES TO ADDRESS

In this paper, we will try to address the challenges facing DNNs/ANNs, as discussed widely by the community [18~22, 25]. In particular, we will consider the observed facts on biological intelligence, such as random interconnection, short response latency, and fault-tolerance, etc. In summary, we will address several open questions detailed below and find the ground for our further analysis.

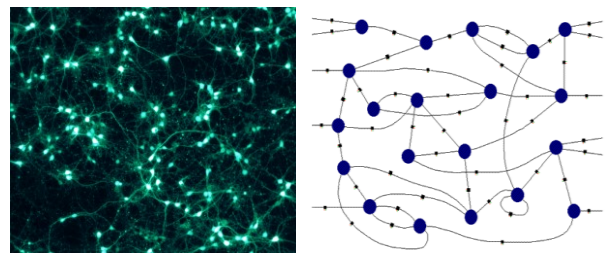

Figure 2. The real random connectivity from a fluorescence microscope [52] of a human brain (left) and the Turing's unorganized neural network [26] (right).

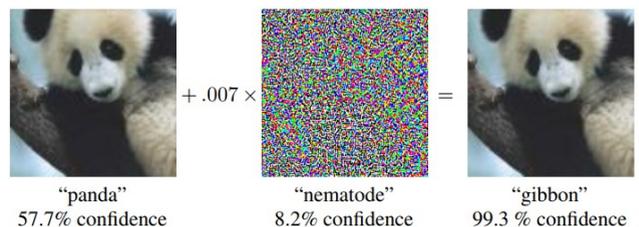

Figure 3. A minor noise attack can easily trick DNN-based classifiers to mistake a panda as a "gibbon" [20].

1) *Can a single neuron have its independent intelligence?*
The common statement that ANNs are inspired by the neural structure of brains is only partially true. Cells in artificial networks such as MLPs (multilayer perceptrons) or RNN (Recurrent neural networks) are not like cells in brain networks.

A neuron cell in ANNs and DNNs, once its weights are trained, will simply act like a fixed weighted sum calculator of input signals with no adaptable capability to discriminate a signal from others. Actually, the intelligent functions of ANNs and DNNs rely mostly on having well-designed architectures. An unorganized architecture cannot work in the way ANNs/DNNs do.

In biological observation, it has been found that a single biological cell can manoeuvre over complex tasks and respond to stimuli in very sophisticated ways [39]. Obviously, the behaviour of a single cell is much more complicated than the simplified weighted sum model in ANNs.

ANNs may be considered as a case of Collective Intelligence (CI) [29], while each neuron is the same unit but configured to different roles. From the viewpoint of collective intelligence [29, 28], if a single unit has zero intelligence, the total intelligence is zero as well. Therefore, collectively we expect a single neuron can cope with discriminative tasks over signals (inputs). To address this view of collective intelligence and finally lead to the understanding of the biological intelligence, a new artificial neuron model is needed to model a different mechanism of computational intelligence.

2) *Can unorganized architectures work?*
Neural Nets were vaguely inspired by the connections observed between the neurons of a brain. Initially, there probably was an intention to develop ANNs to approximate biological brains. However, modern ANNs are not designed to provide a functional model of an animal brain.

Nearly all types of DNNs need a clear definition of interconnections between neurons. For example, in CNNs, convolutional filters need to be applied to neighbouring pixels, and hence a neuron in the current layer need to be carefully connected to the specified neighbouring neurons in the previous layer. While fully random interconnections have been predicted by Turing's proposal on type-B unorganized machines, it is not yet clear how such randomly interconnected networks can work in practice.

Actually, we still do not understand how brains learn, or how redundant connections store and recall information. Brain fibres grow and reach out to connect to other neurons, neuroplasticity allows new connections to be created or areas to move and change function, and synapses may strengthen or weaken based on their importance. "Neurons that fire together, wire together", as Hebb suggested [31].

3) *Capability of fault-tolerance*
Biological neural networks, due to their topology, are also fault-tolerant. Information is stored redundantly so minor failures will not result in memory loss. Neurons in the human brain do not possess numerical values at double precision, and also synaptic inputs do not have any negative weights. In contrast, ANNs are sensitive to the precision of values at the inputs or the parameters. Slight variation in values can impact their final decisions, as shown in Fig.3 [19, 20].

Biological intelligence comes from the interconnections between myriads of neurons in the brain [25]. Adaptability has been a fundamental feature of a biological neural system. Removal of one or several neurons from human perception system could have little impact on the final decision made by humans, while DNNs/ANNs cannot afford any changes in its carefully designed and fine-tuned architectures [19].

4) *Deep or shallow?*
Thorpe *et al* [23] found that monkeys and humans were able to detect the presence of animals in a visual scene in an extremely short time (~100*ms*), leaving neurons just enough time to fire a single spike. Thorpe *et al* suggested that the phenomenal amount of computation achieved in a short time by the human visual system is clearly a challenge for current theories of neural computing such as DNNs. It contradicts the assumption that visual computation is carried out via many-layer propagation over neural networks as DNNs do. Instead, it restricts biological neural systems to a very shallow architecture with 1-2 layers of propagation.

DNNs can have hundreds or even thousands of layers in their architecture, and information in artificial neurons is carried as continuous, floating point values of synaptic weights. The success of DNNs, on the other side, leaves BNNs with the puzzle of how a shallow, randomly connected, parallel biological neural system can solve challenging perceptual tasks.

### C. MOTIVATION OF OUR WORK
The main purpose of our work is to find a practical workable mechanism of unorganized neural networks, namely Turing's B-type machine. To achieve this, we propose a new bio-plausible computational neuron model for computational purposes, and we develop practical unorganized architectures for machine learning tasks.

Our new model (namely SUNNs) will have three features which are distinct from standard ANNs and DNNs:
1) SUNNs are based on a new artificial neuron model, namely *Smart Neuron Model*, which has its own intelligence to selectively process the input signals.
2) SUNNs have a brain-like random-interconnected topology, making them a practical demo of Turing's B-type unorganized machine.
3) Rather than relying on "deep" architectures, SUNNs can have very shallow architectures. In this paper, we present a dual-layer architecture that needs only one spike propagation.

In the experimental validation, we take some early-stage vision tasks as our test bed. Surprisingly, our experiments show that unorganized networks can work nicely on a number of visual tasks. The success of our new computational model makes it clear for the first time that an unorganized network (Turing's B-Type machine) can work as well as (or better than) a well-defined network.

## III. UNORGANIZED NEURAL NETWORK

### A. RANDOM INTERCONNECTION IN HUMAN BRAIN

From biophysical experiments, it is clear that biological neural circuitry has great freedom to interconnect randomly, as shown in Fig.1 [18, 19, 25, 26]. Turing has advocated such random networks and coined the concept of type-B unorganized machine. However, it could be argued that, though such random interconnections do happen in the brain, it may not be efficient, and hence a carefully designed architecture could be more computationally efficient for machine learning.

Interconnections in the brain are not universally random. Recent experiments [32, 33] show that the brain is functionally partitioned, and random interconnections are more likely to be restricted to a local range rather than globally. Mostly, interconnected neurons may cope with similar functions, or in a cascaded pipeline of signals.

Other than believing that such randomness is a waste of resources, is there any deeper reason for such randomness in terms of the efficiency of computing? Is Turing's B-type unorganized machine obsolete or are we on the way to getting closer to the true nature of human intelligence?

To answer these puzzling questions, we would come back to McCarthy's comments *"As soon as it works, no one calls it AI anymore."* [28] The straightforward way to prove Turing's B-type machine is to find a workable implementation of such a randomly interconnected neural network and prove it is efficient. While this is the motivation of our work, before we move into the practical, we start from a tentative theoretical model for the randomness in unorganized networks.

### B. MODELLING OF UNNs BASED ON COST FUNCTION

While unorganized networks lack a clearly defined architecture, there is so far no mathematical model to capture its functional basis yet. While all interconnections are randomly generated, data flows and functionality distribution are totally unplanned and beyond prediction. It is yet a puzzle for neuroscientists to understand how an unorganized network, like the one shown in Fig.2, can work properly with no error. As our 1st step, here we will try to propose a tentative model as the starting point for further modelling.

In the paradigm of machine learning, "intelligence" is often interpreted as a capability to classify patterns or fit data with a learned model [34]. Mathematically, it is often modeled as a cost function of a set of patterns:

$$\arg\max G(x \mid X) \quad (1)$$

where $x \in R^N$ is the input data that needs to be evaluated intelligently, $X$ denotes the labeled or unlabeled dataset that can be used to judge on $x$, and $G$ stands for the cost function to make the optimum decision.

Here, $x$ is a vector belonging to the Riemannian space of $N$ dimensions. As advocated in [36], there could be Many Graph Embeddings (MGE) underlying in its subspaces. To find out these discriminative graphs, we can randomly split this $R^N$ space into a set of subspaces via random subspace sampling [35, 36] and try to find out a discriminative data structure in each subspace. Such a MGE framework has shown much better accuracy [36~38]. Hence, we can introduce this MGE process and produce a set of subspaces,

$$\{s_r \sim x[r_s], S_r \sim X[r_s]\} \quad (2)$$

For a neuron in an unorganized neural network, the above random subspace sampling resembles the random interconnections from the input fields, while each neuron just picks up a small set of signals to process.

With the MGE process, the cost function in Eq.(1) could be split into a set of subspaces as well:

$$G(x \mid X) = \sum_r g_r(s_r \mid S_r) \quad (3)$$

While each $g_r$ is a positive item, we can then have:

$$\arg\max G(x \mid X) = \sum_r \arg\max g_r(s) \quad (4)$$

With the above equation, optimizing the global cost function becomes an easier local optimization problem.

In unorganized neural networks (UNNs), the above $g_r$ can stand for an optimization task for a single neuron. As a result, an intelligent task modeled by Eq.(1) is then naturally split into "single cell"-based cost functions, while each neuron can randomly select its input signals $s_r$ from the input data $x$.

As denoted in Eq.(4), each neuron will process a subset of the input signals in an independent way with some sort of "intelligence". Obviously, the conventional weighted-sum calculator model in ANNs cannot cope with this task. We then need a smarter neuron model that can cope with the needs of the UNNs.

As has been reported, random subspace sampling can produce better accuracy with robustness to noise and also higher fault-tolerance. With the above tentative model, we provide a tentative mode to explain how the bio-plausible Turing B-type unorganized networks can have potential advantages over accurately-designed and fine-tuned neural networks such as ANNs/DNNs.

## IV. SMART NEURON MODEL

### A. ASSUMPTIONS FROM BIOLOGICAL FACTS

In the standard ANNs/DNNs, a neuron is modelled simply as a fixed weighted-sum calculator, as shown in Fig.1. This is a bit too simple compared with the real biological system. Also it lacks bio-plausibility in the sense that a biological neuron has no negative weights [25] in its synaptic inputs, and also cannot hold double-precision values in its weights.

In the natural world, even a single-cell organism can maneuver smartly to adapt to the environment and prey on food [39]. From the view of collective intelligence [29, 30, 31], human intelligence could be an automatic integrated outcome of smarter neurons.

Logically, if a neuron can cope with an intelligent task independently, the overall intelligence can be easier to achieve. Taking this further, the interconnections may become less important is most functionality is embedded in each neuron.

Then, what type of intelligent neurons can we expect? As it has been reported [40], a biological neuron does change its synaptic weights over different signals. Hence, weights cannot be fixed values. Instead, they should be *discriminative* as a function of its input signals *s*. Based on this idea, we come up a new computational model of neurons, namely our *Smart Neuron Model* (SNM), as a fundamental building block for our new neural networks.

There are many "spiking" versions of DNNs/ANNs [25, 41, 52, 54-56]. However, these spiking versions mostly have the same or similar computational mechanism as DNNs/ANNs do, such as using convolutional filters and RELUs [55-56]. Fundamentally, their computation were based on the ANN's weight-sum neuron model. In our work, we aim to develop a more bio-plausible ANN model, which is fundamentally different from the standard ANN model.

### B. SMART NEURON MODEL

In the ANN model shown in Fig.1, each artificial neuron is simply a fixed product-sum regardless of what kind of signals/patterns are input:

$$y_k = \sum_j w_{kj} x_j \quad (5)$$

It is simply a calculator with little intelligence using a set of fixed weights, while these weights will not change no matter what kind of patterns/signals the neuron receives.

To remedy this issue in ANNs, in this paper, we propose a Smart Neuron Model (SNM) as a new type of artificial neurons. Based on MGE, assuming each neuron will try to extract a meaningful discriminative pattern from a subspace, its weights will be an optimized result as a response to *x* over a cost function,

$$\varphi \sim \arg\max_\varphi F(\varphi | x) \quad (6)$$

Here, $\varphi_k(x)$ is a discriminative function to produce a set of weights that can enable the *k*-th cell to adaptively respond to the input vector *x*. This makes the intelli-cell much different from the classic neurons: its response relies on the current input signals/patterns, rather than just adding them up and pass the sum through like neurons in ANNs or DNNs.

For example, using a two-class Gaussian model, $\varphi(x)$ will split all input signals $x_j$ into two classes and let one type of signals pass and block another type of signals. Then the above cost function can be the ratio of the inter-class distance divided by the intra-class distance, and its best solution $\varphi(x)$ can then be a Gaussian mixture model.

The output from such a smart neuron will then consist two parts, the intensity *v* and the vector of a scan over $\varphi(x)$ as its identified "pattern". The intensity can be estimated as,

$$v_k^{out} = \sum_i \varphi_{k,i} v_i^{in} \quad (7)$$

Here, the input $v^{in}$ may not be necessarily associated with the pattern signal *x,* or simply an average of the input signals at the synaptic input. The identified pattern will be expressed as a train or a vector of values. For example, we can simply convert $\varphi(x)$ into a binary vector using an adaptive threshold.

It is noted that $\varphi(x)$ can be designed in a wide range of forms to stand for different types of neurons. It could be an unsupervised clustering function (such as Gaussian Mixture model), to split the input signals into two or more clusters. Alternatively, it could be an incremental dimensionality reduction processor to learn embedded graphs [36, 37] from data flow and project the input signals onto a favored dimension, either linearly or non-linearly.

Biologically, $\varphi(x)$ may denote the preference of a neuron to respond to a favorite pattern. For example, human eyes are likely to choose green color more than others. Hence, weights to green colors could receive a biased value and a neuron could act like a green color filter. Of course, human neurons could have more complicated mechanisms to choose their favored response to signals.

## V. DUAL-LAYER SHALLOW UNNS

### A. 2D DUAL-LAYER ARCHITECTURE

Since the spiking latency of a neuron can take over 100ms, human perception systems need to respond or make a decision in at most 2-3 spiking propagations. Surprisingly, our human perception systems can perform various complex perception tasks in the short time that points to very shallow propagation of signals. Inspired by this neurologic fact, instead of going "deeper", in this work we will attempt to design a shallow UNN architecture, specifically for early-stage vision tasks.

In our proposed architecture, we assume there is an input layer holding the input signals and a processing layer to analyze the input patterns/signals. Fig.5 shows the proposed architecture. In Fig.4-a), neurons in two layers are paired accordingly to withhold the same signals from image pixels. Assuming two paired neurons are close enough to each other, they both share the holding of the values of a pixel. The processing layer will then analyze and adjudicate on these pixels via each corresponding neuron.

Apart from the paired interconnection, the other interconnections from a neuron can be randomly generated, see Fig.4-b). Here, a set of random numbers $\{r_x, r_y\}$ are generated within a local region,

$$\{r_x, r_y\} = rand() \times 2R - R \quad (8)$$

rand() will generate two random numbers between 0 and 1, and the above equation will produce a random location within a square region of *R* (namely from –*R* to +*R*). With the 2D net in Fig.5, we can easily apply the UNNs to 2D signal processing such as image analysis. In our experiments, we choose 8*R* random connections per neuron.

If we stack the paired neurons $k^{in}$ and $k^p$ together, we can have a simplified overview of the shallow dual-layer architecture, as shown in Fig.4-c). It is exactly a simple example of Turing's unorganized machine in Fig.2.

For dual-layer UNNs in this paper, we use a simple Gaussian function to construct $\varphi(s)$,

$$\varphi(s) = N(s_k, s_i) \quad (9)$$

Here *N* is a Gaussian function to estimate the likelihood between signal $s_i$ and $s_k$. Being more similar to each other, their neurons are more likely to connect together, with a larger weight. Somehow the function $\varphi$ serves as a gate to control the signal flows and divide the input signals into two groups: similar to $s_k$ or not.

Note that our Smart Neuron Model is not limited to the above Gaussian function based simple implementation, which is based on finding similarities among all inputs/signals. In the longer term, we could envisage a set of possible functions with automatic selection of the "most responsive" function based on its input data. But for now, in this initial paper, we leave this issue open, and we focus on using specific hand-crafted functions, in this case the simple model based on Eq.(9).

From Eq.(9), a neuron will have a train of weights $\varphi$ that denote the connectivity from the current neuron to its surrounding neighbors. To capture the connectivity map, we can use the average of $\varphi$ to describe the degree of connectivity to its local neighbors:

$$\widetilde{\varphi} = \frac{1}{K} \sum_k \varphi_k(s_k) \quad (10)$$

Here, $\varphi_k$ is the weighted factor to the *k*-th input $s_k$. From the above equation, we will obtain a connectivity map (c-map) of all neurons (or pixel in this specific case).

Fig.6-b) demonstrates such a c-map, which seems more like a sketch conversion of the original image in Fig.6-a), and it captured the most noticeable edges with hierarchical emphasis via its weights.

### B. MODELLING THE LEAKY PROCESS IN UNNS

Recent research on DNNs [38, 39] has suggested that synaptic leaky current in neurons may play an important role in learning, such as contributing to a bio-plausible mechanism for back-propagation. In our work, we further consider this in our model of shallow dual-layer UNNs.

As shown in Fig.5, a 2D net is constructed with the interconnection weights computed from $\varphi$. Considering each neuron can release its voltage potential to its connected neighbors, we can run a leaky test by assuming all neurons have the same potential at the very beginning, and simulate how the potential could be drained out via the 2D net.

In the leaky process, for each neuron, it will release its potential to its connected neighbors at each step:

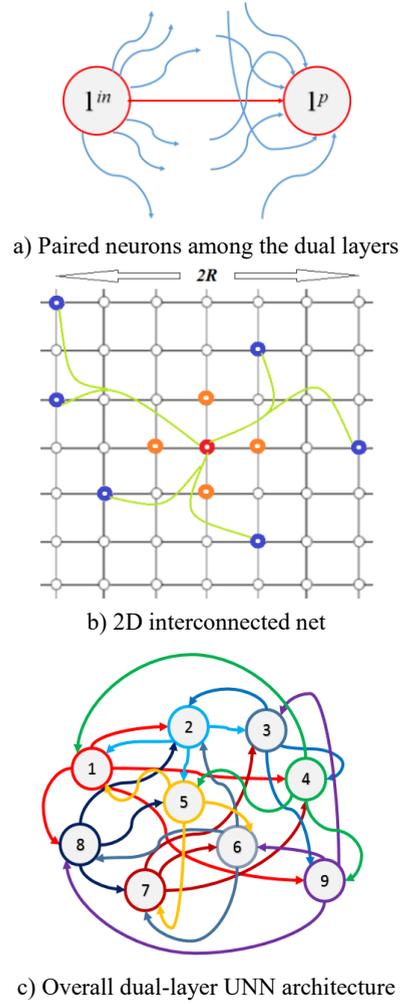

a) Paired neurons among the dual layers

b) 2D interconnected net

c) Overall dual-layer UNN architecture

Figure 4. The proposed shallow UNN architecture

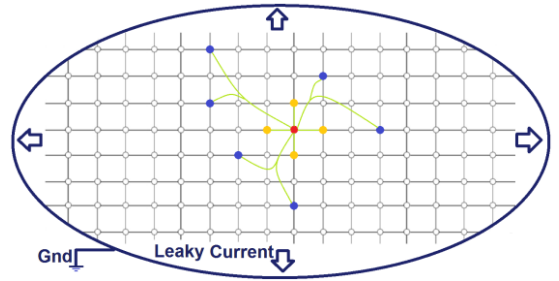

Figure 5. Modelling visual perception using 2D dual-layer UNN. With the connectivity in UNNs, each neuron (shown as the red node) may leak its voltage potential over the net to the surrounding ground.

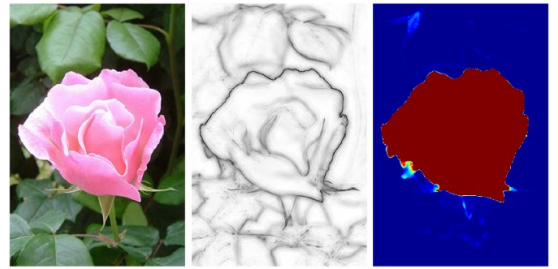

Fig.6 Output results from SUNN. a) Image; b) c-map; c) PR map. Here, the greyscale PR map is illustrated by the 'jet' colour map.

$$\Delta v_k^{out} = \sum_i \varphi_{k,i} \Delta v_i^{in} \qquad (11)$$

In our leaky test, we assume all neurons have the same starting potential, and then the iterative process is applied to see how fast the potential $v_k$ of the neuron $k$ can be released via the 2D net with the setup of $\varphi$.

This leaky test can hopefully examine the interconnectivity between neurons in the 2D net, and as a result analyze the structure of input 2D signals, ideally an image. This serves our purpose for visual perception. Fig.6-c) illustrates the outcome of the potential residue (PR) map by applying the iterative leaky process in Eq.(11) to the sample image in Fig.6-a). To illustrate the iterative process, Fig.7 shows another example, while the PR map is successively extracted by the iteration process.

As far as the leaky process is concerned, the computation of the neural network need an iterative process, where a neuron's output relies on its interconnection with all neighbors and indirectly the whole network. However, from our experiments, this iterative process can quickly converge in 10-20 iterations. This also means that the corresponding implementation of the above computational neural network on hardware or circuits has to be asynchronous, while the outputs may need to wait for the "popping-out" from accumulation as the real biological neurons do.

## VI. USING SUNN FOR EARLY-STAGE VISION TASKS

The human perception system mostly processes 2D signals. Even auditory systems convert 1D audio into 2D spectrum response for further recognition. Our brains process both sight and sound in the same 2D manner, as found by neuroscientists [43]. Hence, our 2D dual-layer SUNNs may match these other perceptual tasks. Here we take visual perception as our test domain.

Visual perception, particularly early-stage vision, begins with the task of how to segment an image and find out what relevant content or objects are in an image. First of all, detecting primary edges are considered the fundamental step for any further analysis. In our experiments, we demonstrate how our SUNNs can cope with edge extraction, and then show a potential use of SUNNs for objectness estimation.

### A. EDGE EXTRACTION

Edge detection is the most important step for human perception to understand the structure and contents of an image or a visual field. It has been shown in neuroscience that our human vision system processes edge extraction at its early vision stage, before any further high-level processing [20, 21]. Hence, we take edge detection as our first experiment to demonstrate how SUNNs could work on real visual tasks.

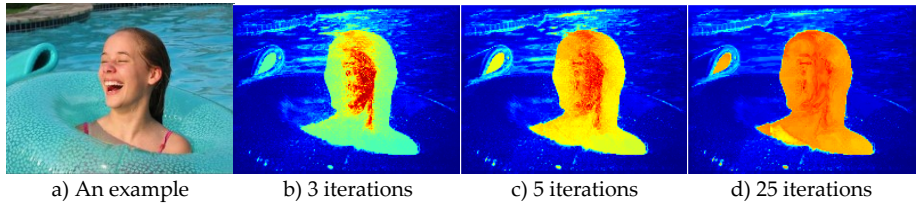

a) An example    b) 3 iterations    c) 5 iterations    d) 25 iterations

Fig.7 Iterative results of the potential residue map.

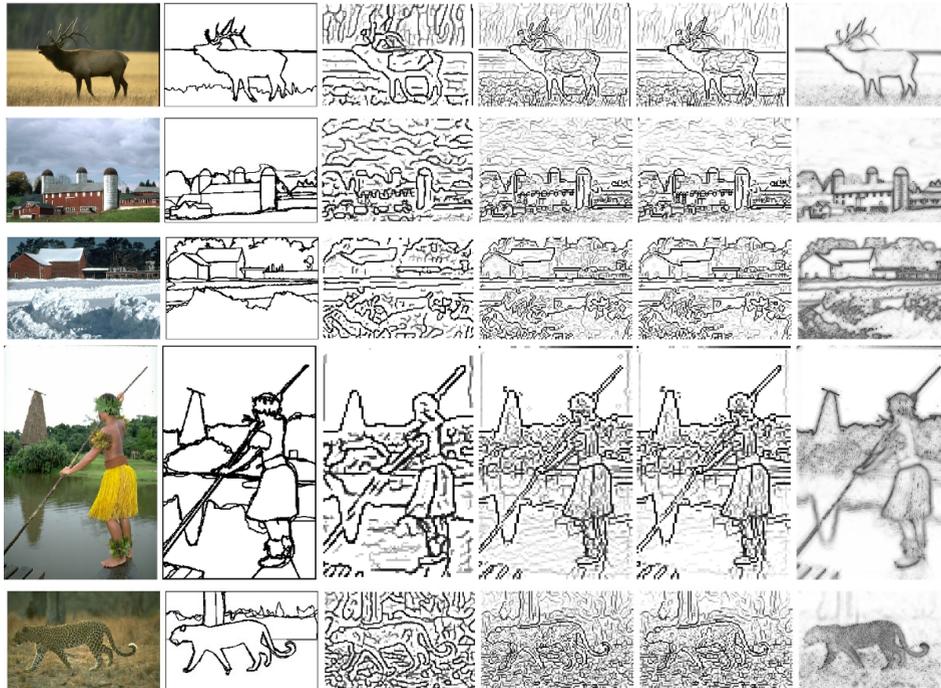

Fig.8. Unsupervised edge extraction. From the left: image, manually labelled, and edges detected by Canny, Sobel, Prewit and our c-map.

In the dual-layer SUNNs described in Section V, the synaptic weight function $\varphi$ stands for the interconnection between neurons, while each neuron holds the value of a pixel. Therefore, we can estimate if a neuron and its corresponding pixel can be on an edge by averaging its synaptic weights, as described in Eq.(10). As a result, we can obtain a connectivity map of the image, which is a match of the segmentation edges of the structure of an image.

Fig.8 shows the edge detection results using our dual-layer SUNNs, in comparison with several classical edge detectors as well as human labelling. Here, Canny, Sobel and Prewitt edge detectors [41, 42] are all based on neighbor-to-neighbor convolutions, the fundamental operation in convolutional neural networks.

From the qualitative comparison, we can see that the edge maps extracted by classical edge detectors contain many local chaotic patterns, while the DL-SUNN based edge maps are more like a sketch abstraction from the original images. Simply by looking into the sketch-like edge map from DL-SUNNs, our human eyes can still easily understand the contents of images, such as deer, houses, person and leopard in the sample images. Obviously, the map extracted by DL-SUNNs is a better match with human perception.

Fig.9 shows the statistical results of edge detection for comparison on the well-known CalTech image segmentation dataset. We can see from the range of recall rates that the DL-SUNN based edge maps achieved the best match with the manually labeled results.

It is worth noting that our purpose is not to provide a better edge extraction algorithm for image segmentation. There are numerous image segmentation methods that can extract edge maps. However, these methods are mostly supervised, while our method is unsupervised. The edge maps coming from DL-SUNNs are natural responses from randomly-interconnected smart neurons to input visual signals, making it meaningful to understand the mechanism of human vision, which is a key purpose of this work.

While we have successfully demonstrated how randomly-interconnected SUNNs can work on edge extraction with no carefully-defined architecture, we now move to a more intriguing topic: the awareness of objects by neurons. Usually, if we can correctly segment an image, it is not far from finding which regions belong to the same semantically meaningful objects.

### B. OBJECT POPOUT FROM DUAL-LAYER SUNNS

The leaky current model has been widely suggested recently as a potential key mechanism for bio-plausible neural computing. It has been advocated that such a leaky process could be a match for the backpropagation process. As detailed in section V-B, we propose a computational leaky

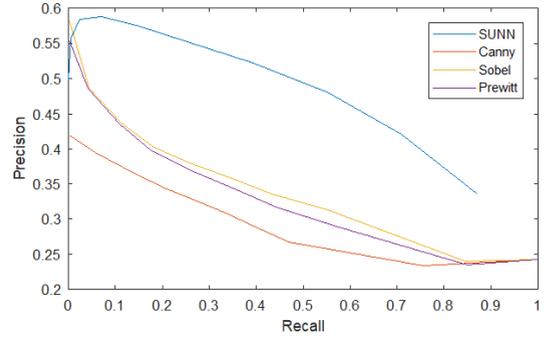
Figure 9. Edge detection on CalTech dataset.

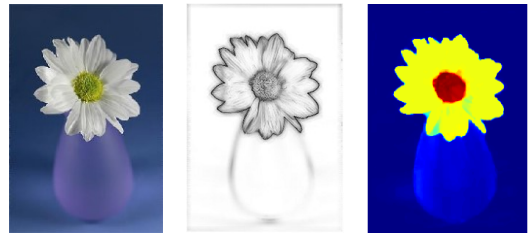
a) From the left: sample image, c map, and PR map.

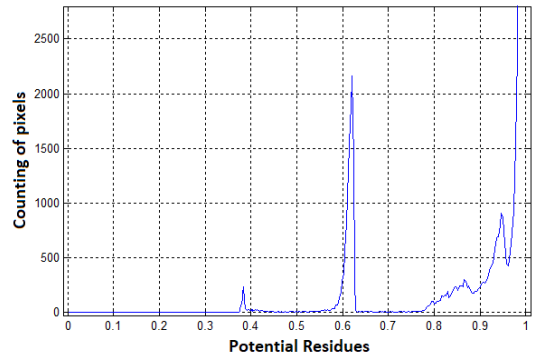
b) Histogram of the PR map.

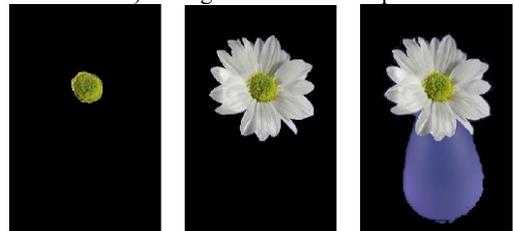
c) Successive pop-out of hiearchical components.
Figure 10. Object pop-out from dual-layer SUNNs.

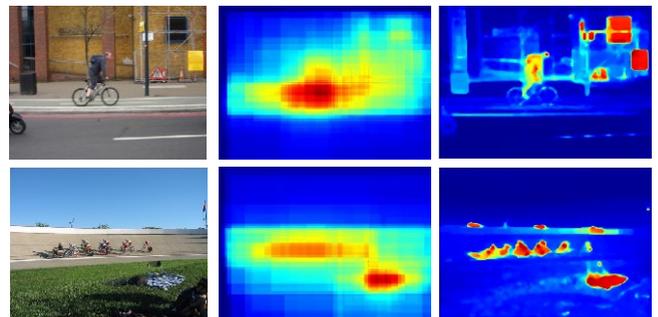
a) Sample images    b) Objectness [42]    c) our PR map
Figure 11. Comparison of objectness estimation.

mechanism jointly with our Smart Neuron Model. After the interconnection $\varphi$ of neurons is set up, we may then iteratively test how voltage potentials, $v_i$, will be leaked out over the net.

Fig.10-a) shows a sample case of such leaky process with our smart neuron model. The left is the sample image, the middle one stands for the interconnection map $\varphi$, and the right one is the iteration result of the potential residue (displayed with the 'jet' color map).

Fig.10-b) shows the historical curve of the potential residue map. The peaks may correspond to its hierarchical structure. If we put a threshold on the residue map, we can easily obtain the results in Fig.10-c), which shows a successive popout of "objects": the pistil, the whole flower, and the vase. Here we come up with a hierarchical concept of objects: an object may contain multiple components, which are objects as well. Hence, the object detection may have not only one correct answer for an image.

In computer vision, many methods were developed to estimate "objectness". Fig.11 shows a comparison of our PR map against a typical objectness estimation method [46]. We can see that our PR map achieved a more accurate map on the objects in the images, such as cyclists, windows, people, etc.

Fig.12 shows more examples of object popout. We see the popped-out objects can be clearly identified with no need for extra segmentation. Note that our method is fully unsupervised. The pop-out of objects is a natural consequence of the leaky process while the produced PR maps were successively thresholded. The experimental results also point to the importance of the leaky process in neural network based computation.

This popout process may have a counterpart in computer vision, namely salient object detection. We combined our dual-layer SUNNs with a center-loss fusion strategy [37] for salient object detection and tested it on the MSRA-1000 dataset. The results are then compared against several well-known unsupervised salient object detection methods, including: 1) LC [47]; 2) CA [48]; 3) HC [49]; 4) FT [50]; 5) RC [49]; and 6) SF [51]. The comparison is made based on the standard recall-accuracy curves, as detailed in [50]. As shown in Fig.13, we can see that our SUNN based method achieved the best results.

The above tests successfully demonstrate that unorganized neural networks, even without a well-defined architecture, can work well for establishing awareness of objects. In computer vision, there are many supervised object detection and segmentation methods which can achieve better results on object detection. For example, Mask-RCNN provided by Facebook Research can detect and segment objects from cluttered backgrounds. However, it is no surprise that a computer can achieve better accuracy than humans, as has been observed in many cases such as playing chess or games. The purpose of our work is not to provide a better computer vision method. Instead we primarily want to demonstrate how neurons in an unorganized network can cope with intelligent tasks.

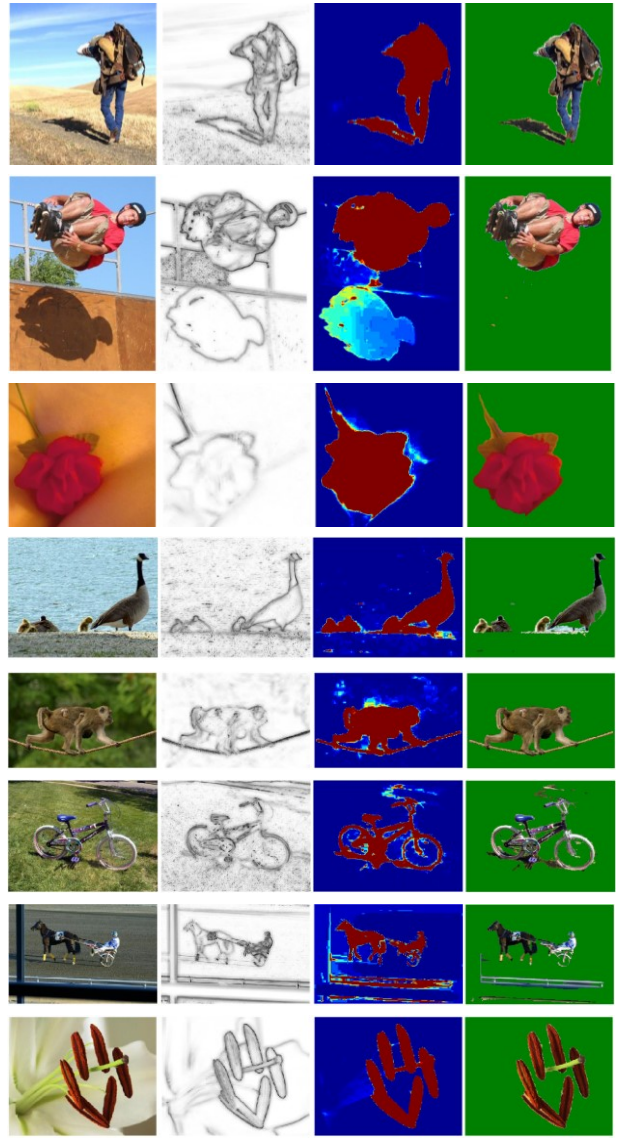

Figure 12. More samples of object popouts. From left: sample images, $i$-maps, PR maps, and the auto popouts.

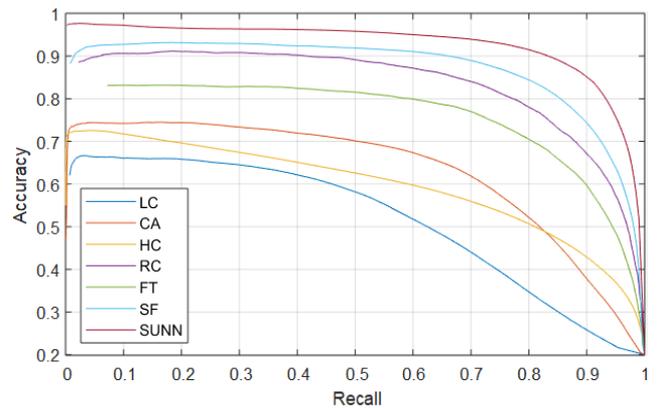

Figure 13. Precision-recall on unsupervised object detection using MSRA-1000 object database [49]. Our SUNN method (top) was compared with six unsupervised salient object detection methods: LC [47], CA [48], HC [49], FT [50], RC [49], and SF [51].

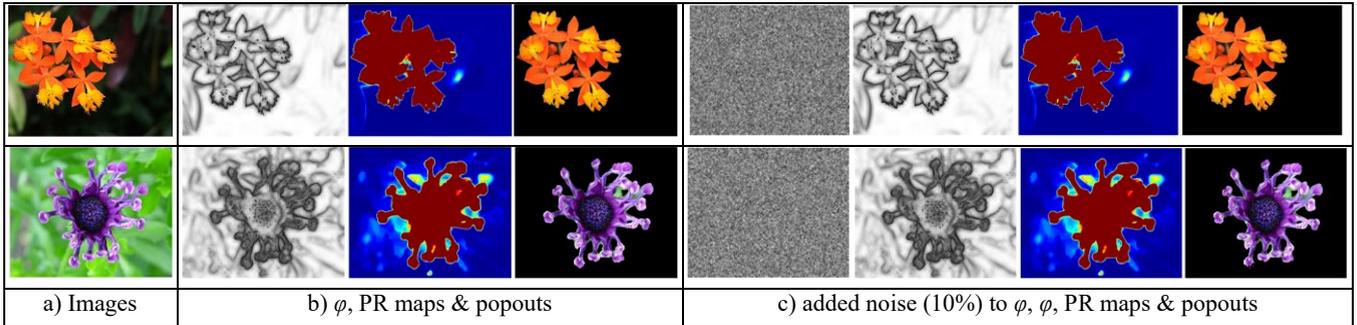

| a) Images | b) φ, PR maps & popouts | c) added noise (10%) to φ, φ, PR maps & popouts |

Figure 14. The synaptic weights can be fuzzy: despite added noise (10%) on φ, similar object popout was obtained.

Fig.14 shows a test with noise attacks. By adding extra noise to φ, we can examine how seriously the added noise impacts the object pop-out process. As shown in Fig.12, we can see that, though considerable noise (10%) is added, the objects were popped out safely. This implies that in our SUNNs, the synaptic weights are not required to be held at double precision accuracy, while ANNs/DNNs cannot tolerate such minor errors. In fact, our human vision system is more like a rough impression-based inference machine, instead of a numerically accurate machine.

## VII. FURTHER DISCUSSION

### A. About Unorganized Machine

Though Turing's B-type machine was proposed nearly 80 years ago [1, 2, 26, 27], almost all ANNs/DNNs so far are based on carefully designed architectures. It is still unknown how randomly interconnected neurons can work together to achieve human intelligence. Our work throws a light on revealing the mechanism by which an unorganized machine can work as well as many state-of-the-art algorithms.

There are arguments that random interconnections bring out the redundancy [41, 52, 19] in biological neural computing, while for machine learning it may not be fully necessary to have such redundancy, and a well-designed architecture could work better at higher accuracy.

In our SUNN model using random sampling, we advocate that such randomness in unorganized networks can not only make it easy to set up the networks (during the growth of nerves), but also help achieve better accuracy with better fault tolerance, as witnessed by many practical reports on random sampling [35-38].

Based on our UNN model, we have proposed a specific network, dual-layer shallow UNN, for early-stage vision tasks. Our experimental results demonstrate that DL-SUNNs can achieve comparable outputs, compared with state-of-the-art methods, in both edge detection, objectness estimation, and unsupervised object detection. Our SUNNs are the first workable Turing's type-B unorganized machine so far. Here 'workable' means it can achieve better or similar performance compared with state-of-the-art algorithms.

### B. Pros and Cons from SUNNs

Our SUNN model provides a more rational bio-plausible computing mechanism. The synaptic weights φ, as shown in Eq.(9), can always be non-negative, unlike the typical ANNs that allow negative weights for convolutional filtering. It also doesn't matter if φ is accurately set up at double precision or not. As shown in Fig.14, added noises does not make any obvious changes in the output results.

Our SUNN model also provides a rational explanation to Thorpe's observation on instant human response to stimuli, while there is only one spike from the stimuli inputs to the object popout. Such a shallow structure may give more space for parallel processing of the input signals, while deeper architecture may not allow too many parallel units due to the limit of resources.

It is worth to note that the dual-layer SUNNs are not the only available model of UNNs. For example, the smart neuron model is not restricted to a simple Gaussian mode in Eq.(9). It can be replaced by any other classifiers, clustering methods, or any other "smart" functions that minimize the local costs in Eq.(6). In fact, in human brain, there are many types of neurons as well, with different functionality. Our dual-layer model may match with the neural computing in retina or primary visual cortex that handle with early vision.

Besides, such unsupervised model can also easily handle with some interesting tasks such as paleographic manuscript recovery [53]. Fig.15 shows the results on separating the paleographic page into two layers using our SUNNs. By running our SUNN model on it, a clear PR map is easily obtained, as shown in Fig.15-b), and then the writings were "popped out" perfectly from its smeared background, , as shown in Fig.15-c). While automated paleographic analysis [53] is yet a very challenging tasks, it is a surprise to see how easily a dual-layer SUNN can cope with it nicely.

### C. Disclaims

As stated in Section I, our SUNN model is not a biophysical model like Spiking Neural Networks. Instead, it is simply another type of ANNs for pure computational purpose, though its basic mechanism is different from the standard ANNs and includes more bio-plausible aspects in its smart neuron model as well as its randomly-interconnected architecture and non-negative weights.

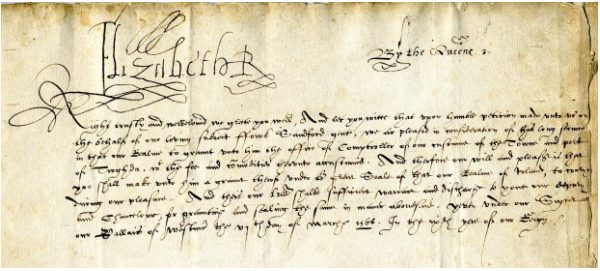
a) the original Shakespear's letter to Queen Elizabeth I

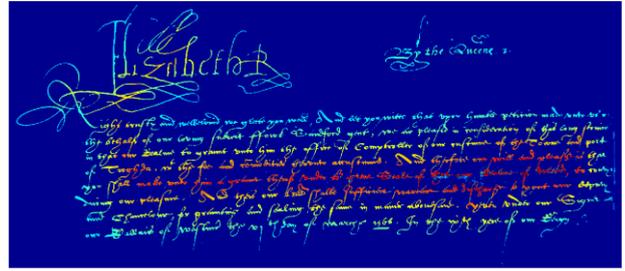
b) the PR map

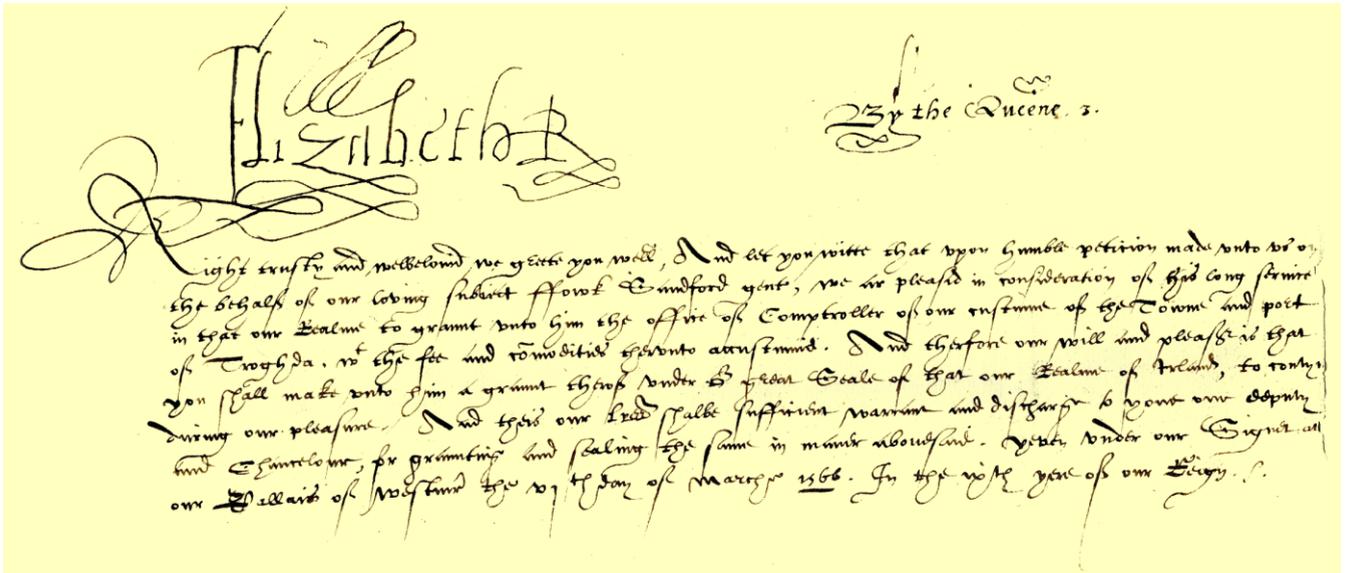
c) Bilayer segmentation results. The background is replace by clear yellow color.

Figure 15. A possible application to paleographic manuscript recovery.

There are many "spiking" versions of DNNs/ANNs [25, 41, 52, 54-56]. They differ from ANNs/DNNs typically in the sense that they simulate the spiking activities to model the biophysical activities of neurons and convert the real-valued inputs or outputs in ANNs/DNNs into spike trains [54]. However, these spiking versions mostly have the same or similar computational mechanism as DNNs/ANNs do, such as using convolutional filters and RELUs [55-56]. Principally, their computation were based on the ANN's weight-sum neuron model to explain the spiking events and hopefully assist the understanding of biophysical processes of neuron computation.

It is worth to note that though our work shows that SUNNs can work well on a number of tasks, it does not imply that a shallow architecture can work better than deeper network architectures. Instead, well trained at double precision with larger datasets, DNNs nowadays can easily outperform human opponents in game plays [8-11]. From a broader view, our model could be complementary to DNNs in the sense that SUNNs provide a new type of artificial neural networks, which have benefits of being fuzzy to noisy signals like human brains.

### D. ASSUMPTIONS FOR NEUROSCIENCE

In the computational model of our SUNNs, we implicitly established its theoretical ground on two neuroscientific assumptions:

1) The synaptic response to a stimuli is not independent and relies on what kind of stimuli other synapses receive. Namely, a neuron can compare inputs and selectively make decision on how to respond discriminatively to a set of synaptic inputs.
2) Leaky current may play an important role before a neuron fires spikes. It provides an extra mechanism for a neuron to communicate with other neurons in the network. Hence, the pop-out from a neuron relies on the whole network, as a consequence of Eq.(11), rather than simply a straightforward weight-sum result of only its own inputs like ANNs, as described in Eq.(1).

While our work exposes the importance of the leaky mechanism in bio-plausible neural computing, it corresponds to the recent research that suspects the real back-propagation of DNNs could be channeled by the leaky process [41, 52]. Our model gives a verifiable computational model of such a leaky process for neuroscientists to examine and verify.

## VIII. CONCLUSIONS

In conclusion, we proposed a practical model of unorganized neural networks, namely Turing's B-type unorganized machine based a new artificial neuron model – *Smart Neurons*, and demonstrated it successfully in experiments with a specific architecture called shallow UNNs. It shows that randomly interconnected UNNs can achieve reasonably good performance in a number of early vision tasks such as edge detection and object awareness, in comparison with the state-of-the-art computer vision methods.

As the first workable model of Turing's B-type unorganized machine, our work throws lights on emulating brain-like randomly interconnected neurons with a rationalized ground based on *many graph embedding*, and suggests that such random interconnection could be of benefits instead of simply redundancy. Such a rationalized brain-like model can address the needs of developing a more bio-plausible new computing model for artificial neural networks and as the final goal, to demystify the origin of biological intelligence.

Our work also revealed that the leaky process could play an important role in the computational process, as suspected for a while by compute scientists. Actually, the computation over our SUNNs is accomplished directly by the leaky process as described in Eq.(11) before the spiking, via the leaky process, and the outputs from a neuron is a consequence of the iteration over the whole network rather than as a single-neuron calculation on its own inputs.